\documentclass{article} 
\usepackage{iclr2026_conference,times}

\usepackage{amsmath,amsfonts,bm}

\def\eqref#1{equation~\ref{#1}}

\def\1{\bm{1}}

\DeclareMathAlphabet{\mathsfit}{\encodingdefault}{\sfdefault}{m}{sl}
\SetMathAlphabet{\mathsfit}{bold}{\encodingdefault}{\sfdefault}{bx}{n}

\usepackage{hyperref}
\usepackage{xurl}
\usepackage{booktabs}
\usepackage{float}
\usepackage{graphicx}
\usepackage{algorithm}
\usepackage{algpseudocode}

\title{DriftST: One-Step Generative Inference of Spatial Transcriptomics from H\&E Histology}

\author{
Yuhang Yang$^{1,*}$,
Yonggan Bu$^{1,*}$,
Shengyuan Zhou$^{2}$,
Yiming Luo$^{2}$,
Kai Zhang$^{1,\dagger}$\\[4pt]
$^{1}$University of Science and Technology of China, Hefei, China \\
$^{2}$Peking University Cancer Hospital, Beijing, China \\[4pt]
$^{*}$Equal contribution \qquad
$^{\dagger}$Corresponding author: \texttt{kkzhang08@ustc.edu.cn}
}

\iclrfinalcopy 

\begin{document}

\maketitle

\begin{abstract}
Spatial Transcriptomics (ST) measures gene expression while preserving spatial context, but its high cost and low throughput leave public datasets small. Inferring expression directly from widely available Hematoxylin and Eosin (H\&E) stained histology offers a cost-effective alternative.
However, existing approaches face several limitations: i) regression methods over-smooth toward the conditional mean, while generative methods are faithful but need slow multi-step inference; ii) most treat genes as independent and equally important, ignoring inter-gene dependencies and the heterogeneity of gene informativeness; iii) and most are tailored to a single resolution, spot-level or cell-level.
To address these issues, we propose \textbf{DriftST}, a unified framework for inferring spatially resolved gene expression from H\&E images. DriftST builds on a Cellular Drifting generative model that learns a direct drift from a histology-conditioned source to the expression distribution, retaining generative expressiveness while enabling efficient one-step generation. To capture gene structure, we introduce the STransformer, which combines a co-expression attention module for inter-gene dependencies with a gene residual gate for differential gene importance. Operating on a generic gene-panel representation, DriftST applies directly to both spot-level and cell-level data in one framework, and extensive experiments across diverse tissues and platforms show it achieves state-of-the-art performance at both resolutions.
Code is available at \url{https://github.com/yyh030806/DriftST}.
\end{abstract}
\section{Introduction}
\label{main:intro}

Spatial Transcriptomics (ST)~\cite{stahl2016visualization} measures gene expression while preserving the spatial organization of cells within a tissue, revealing how expression varies across regions and microenvironments, a spatial context that neither bulk~\cite{wang2009rna} nor single-cell RNA sequencing~\cite{macosko2015highly} can resolve. This is essential for dissecting processes such as tumor invasion, tissue development, and immune infiltration, where cellular function is tightly coupled to position. Current ST technologies fall into two families: sequencing-based platforms (e.g., 10x Visium~\cite{stahl2016visualization}) profile capture spots tens of micrometers wide that typically span multiple cells, yielding spot-level resolution over a near-transcriptome-wide panel; imaging-based platforms (e.g., MERFISH~\cite{chen2015spatially}, Xenium~\cite{janesick2023high}) report true single-cell resolution by imaging individual transcripts in situ, over a more limited pre-selected panel. Together, these complementary modalities have advanced our ability to map the spatial transcriptional landscape of complex tissues.

However, ST remains costly to deploy, demanding specialized equipment, technical expertise, and long turnaround times at hundreds to thousands of dollars per sample~\cite{stahl2016visualization}, leaving public datasets small and narrow. Hematoxylin and Eosin (H\&E) stained histology images, by contrast, are cheap, fast to acquire, and routine in clinical pathology~\cite{chen2024uni}. Since cell morphology in H\&E is closely tied to cell type, state, and thus the underlying transcriptome, inferring spatially resolved gene expression directly from H\&E has become an appealing direction: a reliable solution would unlock genomics-level analysis from abundant imaging data without costly wet-lab profiling.

A growing body of work has tackled this problem. Early methods cast it as spot-level regression, predicting expression directly from H\&E patches, including ST-Net~\cite{he2020integrating}, HisToGene~\cite{pang2021leveraging}, Hist2ST~\cite{zeng2022spatial}, BLEEP~\cite{xie2023spatially}, TRIPLEX~\cite{chung2024accurate}, and M2ORT~\cite{wang2024m2ort}. More recently, generative formulations such as STEM~\cite{zhu2025diffusion} and GenAR~\cite{ouyang2025genar} model the conditional distribution rather than a point estimate. In parallel, the latest efforts, GHIST~\cite{fu2025ghist} and sCellST~\cite{chadoutaud2026scellst}, target the harder cell-level setting, predicting expression at single-cell rather than coarse multi-cell spot resolution.

While encouraging, these methods share several limitations.
\textbf{i) Predicting paradigm.} Regression methods are efficient with only one forward pass per prediction, but minimizing an expected error drives them toward the conditional mean, yielding over-smoothed profiles that fail to recover true biological heterogeneity across locations~\cite{xie2023spatially, zhu2025diffusion}. Generative methods model the conditional distribution far more faithfully, but diffusion- and autoregressive-based formulations need many iterative refinement steps at inference, making them expensive and slow.
\textbf{ii) Model architecture.} Existing methods largely treat panel genes as independent and interchangeable, overlooking two key properties: genes are not independent, acting in concert through regulatory networks, signaling pathways, and co-expression modules~\cite{barabasi2004network}, so predicting each in isolation discards meaningful structure; and genes are not equally informative, as some are highly variable and discriminative while others are near-constant or noise-dominated, yet current models give all equal capacity.
\textbf{iii) Data applicability.} Most methods are designed for a single resolution, spot-level or cell-level, and do not transfer between the two, limiting generality across ST platforms.

To address these limitations, we propose \textbf{DriftST}, a unified framework for inferring spatially resolved gene expression from H\&E images. As a paradigm, DriftST builds on a \textbf{Cellular Drifting} generative model that learns a direct drift transporting a histology-conditioned source distribution to the expression distribution, preserving generative expressiveness and the one-to-many morphology-transcriptome relationship while enabling one-step generation, thus pairing generative fidelity with regression efficiency. Architecturally, we introduce the \textbf{STransformer}, which models gene structure explicitly: a co-expression attention module propagates inter-gene dependencies, and a gene residual gate adaptively reweights each gene by its informativeness. Operating on a generic gene-panel representation, DriftST applies directly to \textbf{both} spot-level and cell-level data in one framework. Extensive experiments across multiple tissues and platforms show DriftST achieves state-of-the-art (SOTA) performance.

Our main contributions are summarized as follows:
\begin{itemize}
    \item \textbf{A novel modeling paradigm.} We reformulate H\&E-to-ST inference as Cellular Drifting, a generative process that learns a direct drift from histology to transcriptome and produces high-fidelity predictions in a single step, resolving the long-standing trade-off between regression efficiency and generative fidelity.
    \item \textbf{A gene-aware architecture.} We design the STransformer, which treats genes as structured rather than independent variables, combining a co-expression attention module that captures inter-gene dependencies with a gene residual gate that adaptively reweights each gene's contribution by importance.
    \item \textbf{SOTA and general performance.} Within a single framework, DriftST applies directly to both spot-level and cell-level resolutions and achieves SOTA results across a wide range of datasets spanning diverse tissues and sequencing platforms.
\end{itemize}
\section{Related Work}
\label{main:related_work}

\subsection{Spatial Gene Expression Prediction from Histology}
Predicting spatially resolved gene expression from H\&E-stained histology
offers a cost-effective alternative to spatial transcriptomics (ST)
experiments~\citep{he2020integrating, xie2023spatially}.
Early work regressed per-spot expression from CNN features~\citep{he2020integrating},
followed by efforts to incorporate spatial context via
Vision Transformers~\citep{pang2021leveraging} and graph neural
networks~\citep{zeng2022spatial}. BLEEP~\citep{xie2023spatially} reframed the
task as contrastive image--expression alignment, while M2ORT~\citep{wang2024m2ort}
and TRIPLEX~\citep{chung2024accurate} fused multi-scale features across magnifications.
More recent generative approaches employ diffusion~\citep{zhu2025diffusion}
or autoregressive~\citep{ouyang2025genar} modeling, and
sCellST~\citep{chadoutaud2026scellst} and GHIST~\citep{fu2025ghist} further
push prediction to single-cell resolution.
However, these methods universally rely on per-spot regression or likelihood-based
objectives, without exploiting structural information from the prediction
distribution to guide learning.

\subsection{Pathology Foundation Models}

Large-scale self-supervised pretraining has produced powerful visual encoders for computational pathology. CTransPath~\citep{wang2022ctranspath} pioneered pathology-specific pretraining with DINO on TCGA data. UNI~\citep{chen2024uni} and its successor UNI2~\citep{chen2025uni2} scaled DINOv2-based training to over 100 million patches, achieving strong cross-task generalization. Complementarily, CONCH~\citep{lu2024conch} adopted vision-language contrastive learning on pathology image-text pairs, capturing semantic-level information beyond pure morphological features. Pure vision models and vision-language models encode orthogonal information, and their concatenation has been shown to benefit downstream prediction tasks.

\subsection{Drifting-Based Generative Models}

\citet{deng2026drifting} proposed the Drifting Model, a generative paradigm that enables single-step generation by shifting distributional evolution into training. Several concurrent works have since analyzed its theoretical foundations: \citet{lai2026unified}, \citet{turan2026secretly}, and \citet{li2026longshort} connected the drifting field to score matching and flow matching, while \citet{he2026sinkhorn} proposed Sinkhorn-Drifting to improve training stability via entropic optimal transport. However, all existing drifting models operate in unconditional or class-conditional settings and none addresses paired prediction tasks such as spatial gene expression prediction.
\section{Methodology}
\label{main:method}

\subsection{Preliminaries}

\textbf{Notations.} Let $\mathbf{X}^{\mathrm{cnt}} \in \mathbb{N}^{N \times G}$ be the
raw ST count matrix for $N$ spatial locations (multi-cell spots or single cells) across
$G$ genes, and $\mathbf{S} \in \mathbb{R}^{N \times 2}$ their spatial coordinates. We
model the \texttt{log1p}-normalized expression $\mathbf{X} = \log(1 +
\mathbf{X}^{\mathrm{cnt}}) \in \mathbb{R}^{N \times G}$, while the raw counts
$\mathbf{X}^{\mathrm{cnt}}$ are retained for the ZINB likelihood (Sec.~3.2). The H\&E
histology image is $\mathbf{I} \in \mathbb{R}^{H \times W \times 3}$, from which we crop
a localized patch $\mathbf{I}_s \in \mathbb{R}^{P \times P \times 3}$ at location $s$.
Our task is to infer the expression profile $\mathbf{x} \in \mathbb{R}^{G}$ (in
\texttt{log1p} space) from $\mathbf{I}_s$, with $\mathbf{x}^{\mathrm{cnt}} \in
\mathbb{N}^{G}$ its corresponding raw counts.

\begin{figure}[t]
\centering
\includegraphics[width=1\linewidth, keepaspectratio]{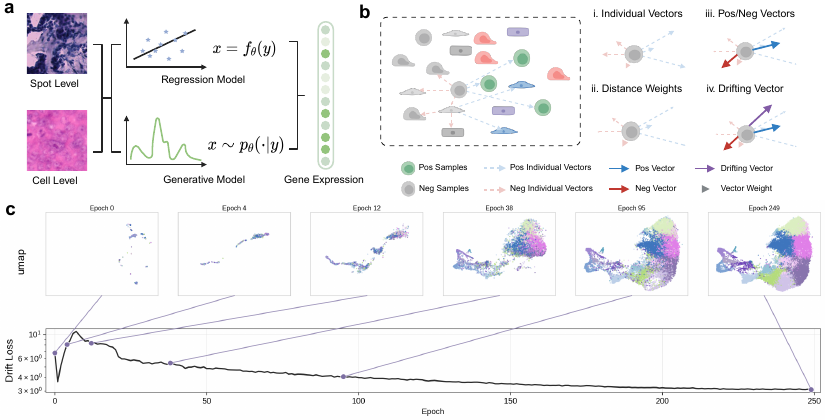}
\caption{\textbf{Framework of the Cellular Drifting.} \textbf{a,} Problem formulation for inferring spatial transcriptomics from spot- and cell-level H\&E images, contrasting regression and generative models. \textbf{b,} The cellular drifting field: positive (ground truth) and negative (generated) samples interact via distance-weighted vectors to form a unified drifting vector. \textbf{c,} As the drift loss decreases over epochs, the generated distribution (UMAP) progressively aligns with the target, enabling one-step generation at inference.}
\label{fig:drifting}
\end{figure}

\begin{figure}[t]
\centering
\includegraphics[width=1\linewidth, keepaspectratio]{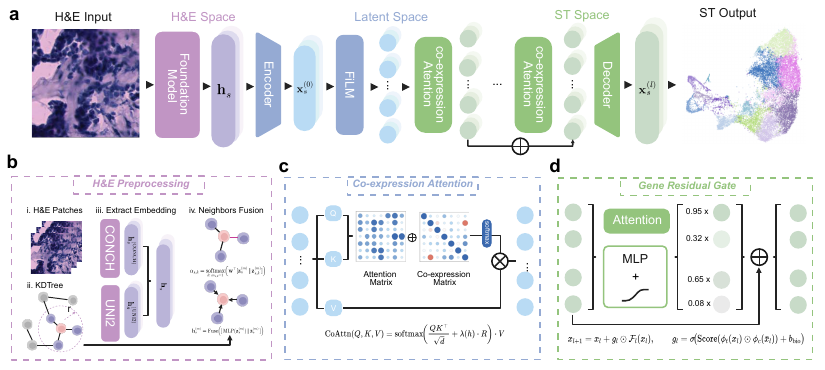}
\caption{\textbf{Architecture of the STransformer.} \textbf{a,} Overall pipeline mapping localized H\&E patches to spatial gene expression through tailored generative modules. \textbf{b,} H\&E preprocessing: dual pathology foundation models (CONCH, UNI2) extract visual embeddings, fused with KDTree-based neighbors for local context. \textbf{c,} Co-expression Attention: self-attention over the gene dimension models biological pathways and co-expression dependencies. \textbf{d,} Gene Residual Gate: adaptive gating reweights each gene by its informativeness and discriminative power.}\label{fig:stransformer}
\end{figure}

\textbf{Overview.} DriftST is a unified generative framework that directly learns the distributional drift from the H\&E morphology distribution $p_{\text{H\&E}}$ to the ST expression distribution $p_{\text{ST}}$, combining the fidelity of generative modeling with the efficiency of a single forward pass. At its core, a Cellular Drifting model (Figure \ref{fig:drifting}) transports the H\&E-conditioned source distribution to the target $p_{\text{ST}}(\mathbf{x})$. By minimizing this drift field during training, the pushforward distribution $q_\theta = [f_\theta(\mathbf{I}_s)]_{\#} p_{\text{H\&E}}$ is driven to match $p_{\text{ST}}$, which removes the need for iterative sampling at inference~\cite{deng2026drifting}. The generator $f_\theta$ is instantiated as the STransformer (Figure \ref{fig:stransformer}), which extracts morphological features from $\mathbf{I}_s$ while explicitly modeling the inter-gene dependencies and differential gene importance of the output space $\mathbb{R}^G$.

\paragraph{Cellular Drifting Field.}
To transport the pushforward distribution $q_\theta$ toward $p_{\mathrm{ST}}$, we
construct a cellular drifting field that decouples the interaction into a positive
attraction and a negative repulsion, which prevents the
single paired target from being overwhelmed by the abundant negatives. For a patch
$I_s$, the generator produces a prediction $\mathbf{x}=f_\theta(I_s)$. We define the
positive set $\mathcal{P}=\{\mathbf{x}^{+}\}$ as the paired ground-truth profile of
$I_s$ (hence $|\mathcal{P}|=1$), and the negative set $\mathcal{N}$ as the union of
the $K{-}1$ other stochastic dropout predictions of the same patch and a pool of
historical predictions sampled from a prediction bank; the prediction's own copy is
excluded from $\mathcal{N}$, so it is never repelled from itself. To stabilize the
vector field in the high-dimensional space $\mathbb{R}^{D}$, we apply a global
distance scaling factor $\sigma_d=\max(\tfrac{1}{\sqrt{D}}\,\mathbb{E}[d],\epsilon)$,
where $\mathbb{E}[d]$ is the mean Euclidean distance from $\mathbf{x}$ to all samples
in $\mathcal{P}\cup\mathcal{N}$, and denote the scaled features as
$\tilde{\mathbf{x}}=\mathbf{x}/\sigma_d$.

We compute the field with a multi-scale formulation over a set of bandwidth ratios
$\mathcal{R}$ (e.g., $\mathcal{R}=\{0.02,0.05,0.2\}$), so that both local and global
structure are captured. For each $r\in\mathcal{R}$, the effective temperature is
$\tau_r=r\sqrt{D}$, and the attraction and repulsion weights are computed
independently via softmax:
\begin{align}
w^{(r)}_{\mathrm{pos}}(\mathbf{x}^{+}) &= \mathrm{softmax}\!\left(-\frac{\|\tilde{\mathbf{x}}-\tilde{\mathbf{x}}^{+}\|^2}{\tau_r}\right),\quad \forall\,\mathbf{x}^{+}\in\mathcal{P},\\
w^{(r)}_{\mathrm{neg}}(\mathbf{x}^{-}) &= \mathrm{softmax}\!\left(-\frac{\|\tilde{\mathbf{x}}-\tilde{\mathbf{x}}^{-}\|^2}{\tau_r}\right),\quad \forall\,\mathbf{x}^{-}\in\mathcal{N}.
\end{align}
By enforcing $\sum w_{\mathrm{pos}}=1$ and $\sum w_{\mathrm{neg}}=1$ separately, we
guarantee that the positive target (a single paired ground-truth sample,
$N_{\mathrm{pos}}=1$) exerts a normalized driving force that strictly guides the
prediction, avoiding numerical dominance by the large negative bank. The velocity
field for scale $r$ is the difference between the expected positive and negative
directions, and the final field averages over all scales:
\begin{equation}
V_r(\mathbf{x})=\sum_{\mathbf{x}^{+}\in\mathcal{P}}w^{(r)}_{\mathrm{pos}}(\mathbf{x}^{+})\,\tilde{\mathbf{x}}^{+}-\sum_{\mathbf{x}^{-}\in\mathcal{N}}w^{(r)}_{\mathrm{neg}}(\mathbf{x}^{-})\,\tilde{\mathbf{x}}^{-},\qquad
V(\mathbf{x})=\frac{1}{|\mathcal{R}|}\sum_{r\in\mathcal{R}}V_r(\mathbf{x}).
\end{equation}

\textbf{Drifting Loss Optimization.} With the multi-scale, split-form velocity field $V(\mathbf{x})$ established, we cast training as a self-consistent iterative optimization: rather than optimizing over a static error surface, the network updates its predictions toward an evolving target guided by the biological drifting field. For a current prediction $\mathbf{x}$ and its scaled counterpart $\tilde{\mathbf{x}}$, we define the local goal state as an incremental drift step with step-size hyperparameter $\eta$:
$$\tilde{\mathbf{x}}_{\text{goal}} = \text{stopgrad}(\tilde{\mathbf{x}} + \eta V(\mathbf{x}))$$
The stopgrad operator is crucial here: by freezing the target coordinate, it prevents the network from minimizing the objective through representation collapse or trivial identity shortcuts. The generator $f_\theta$ is then optimized to contract the distance between its prediction and this stabilized goal, using an MSE drift loss in the scaled feature space:
$$\mathcal{L}_{\text{drift}} = \mathbb{E}_{\mathbf{I}_s \sim p_{\text{H\&E}}} \left[ || \tilde{\mathbf{x}} - \tilde{\mathbf{x}}_{\text{goal}} ||_2^2 \right]$$
Minimizing $\mathcal{L}_{\text{drift}}$ implicitly establishes a contraction mapping that progressively drives the velocity field toward an equilibrium where $V(\mathbf{x}) \to 0$. As shown in Figure \ref{fig:drifting}c, this objective reshapes the pushforward distribution over successive epochs until it mirrors the true transcriptomic landscape. Consequently, iterative sampling and stochastic path integration are entirely bypassed at inference: DriftST needs only a single deterministic forward pass to generate high-fidelity spatial expression profiles.

\subsection{STransformer Neural Network}

To serve as the generator $f_\theta$ within the cellular drifting framework, we introduce the STransformer (Figure \ref{fig:stransformer}a). Unlike standard architectures that treat output dimensions as independent, it is explicitly designed to capture the inter-dependencies and heterogeneous informativeness inherent to spatial transcriptomics. The network first encodes a patch and its microenvironment into a unified spot representation, then unfolds it into gene-specific tokens that are progressively refined by co-expression attention and gene-wise gating before being decoded into the final expression profile.

\textbf{H\&E Preprocessing.} As shown in Figure \ref{fig:stransformer}b, for a patch $\mathbf{I}_s$ we extract two complementary visual embeddings using dual pathology foundation models, CONCH~\cite{lu2024conch} and UNI2~\cite{chen2025uni2}, denoted $\mathbf{z}_s^{(m)}$ for $m \in \{\text{CONCH}, \text{UNI2}\}$, which together capture fine-grained morphological and high-level semantic features. To incorporate the surrounding microenvironment, we build a KDTree over the spatial coordinates $\mathbf{S}$ and retrieve, for each location, the set of neighbors $\mathcal{K}_s$ falling within a radius $r$, with a validity mask $m_{s,k} \in \{0,1\}$. Rather than averaging neighbors uniformly, we aggregate them with a learned spatial attention that weights each neighbor by its relevance to the center patch:
\begin{equation}
\alpha_{s,k} = \operatorname*{softmax}_{k:\, m_{s,k}=1}\!\left( \mathbf{w}^\top [\mathbf{z}_s^{(m)} \,\|\, \mathbf{z}_{s,k}^{(m)}] \right), \qquad
\bar{\mathbf{z}}_s^{(m)} = \text{MLP}\!\Big( \sum_{k \in \mathcal{K}_s} \alpha_{s,k}\, \mathbf{z}_{s,k}^{(m)} \Big)
\end{equation}
where $\mathbf{w}$ is a learnable scoring vector and $[\cdot\,\|\,\cdot]$ denotes concatenation. The aggregated neighborhood context $\bar{\mathbf{z}}_s^{(m)}$ is then fused with the center embedding to form a niche-aware per-model representation, and the two foundation-model streams are combined by an MLP encoder into a single spot representation $\mathbf{h}_s \in \mathbb{R}^D$:
\begin{equation}
\mathbf{h}_s^{(m)} = \text{Fuse}\!\left( [\,\text{MLP}(\mathbf{z}_s^{(m)}) \,\|\, \bar{\mathbf{z}}_s^{(m)}\,] \right), \qquad
\mathbf{h}_s = \text{MLP}\!\left( [\,\mathbf{h}_s^{(\text{CONCH})} \,\|\, \mathbf{h}_s^{(\text{UNI2})}\,] \right)
\end{equation}
Rather than treating genes homogeneously, which risks collapsing predictions into a generic profile, we initialize a learnable gene embedding matrix $\mathbf{E} \in \mathbb{R}^{G \times D}$ to modulate $\mathbf{h}_s$ via Feature-wise Linear Modulation (FiLM)~\cite{perez2018film}. The initial hidden token for gene $i$ is
\begin{equation}
\mathbf{x}_i^{(0)} = \text{Proj}\!\left( \gamma(\mathbf{e}_i) \odot \mathbf{h}_s + \beta(\mathbf{e}_i) \right),
\end{equation}
which disentangles the shared histology into gene-specific tokens $\mathbf{X}^{(0)} \in \mathbb{R}^{G \times D}$, so the model can capture the distinct morphological cues required for each gene.

\textbf{Co-expression Attention.} As shown in Figure \ref{fig:stransformer}c, we model inter-gene regulatory dependencies through a sequence of bio-guided self-attention blocks operating along the gene dimension. In each layer $l$, the input tokens $\mathbf{X}^{(l)}$ are first normalized via Adaptive Layer Normalization (AdaLN)~\citep{dhariwal2021diffusion} conditioned on $\mathbf{h}_s$, so that morphological context continuously guides feature refinement:
\begin{equation}
\text{AdaLN}(\mathbf{X}^{(l)}, \mathbf{h}_s) = \gamma_l(\mathbf{h}_s) \odot \text{LayerNorm}(\mathbf{X}^{(l)}) + \beta_l(\mathbf{h}_s),
\end{equation}
where $\gamma_l(\cdot)$ and $\beta_l(\cdot)$ are layer-wise scale and shift parameters predicted from the spot representation. To inject structural biological priors, we add a global gene co-expression matrix $\mathbf{R} \in \mathbb{R}^{G \times G}$ built from the training set as an additive attention bias:
\begin{equation}
\text{CoAttn}(\mathbf{Q}, \mathbf{K}, \mathbf{V}) = \text{softmax}\left(\frac{\mathbf{Q} \mathbf{K}^T}{\sqrt{d}} + \lambda(\mathbf{h}_s) \cdot \mathbf{R} \right) \mathbf{V},
\end{equation}
where the query, key and value are linear projections of the normalized tokens, and $\lambda(\mathbf{h}_s) \in \mathbb{R}^H$ is a spot-dependent, per-head scaling factor predicted from $\mathbf{h}_s$. This allows the model to calibrate its reliance on co-expression priors according to the local tissue state, maintaining strict regulatory rules in normal regions while relaxing them to accommodate pathway reprogramming in anomalous tumor microenvironments.

\textbf{Gene Residual Gate.} Since the transcriptomic landscape is highly uneven, with stable housekeeping genes requiring minimal processing compared to complex biomarker genes, we introduce a progressive Gene Residual Gate (Figure \ref{fig:stransformer}d). For each gene token at layer $l$, a soft gate $g_l \in (0, 1)$ is dynamically computed from its current state, the global context $\mathbf{h}_s$, and its connectivity in the co-expression graph $\mathbf{R}$:
\begin{equation}
g_l = \sigma\!\left( \text{Score}\!\left( \phi_t(\mathbf{X}^{(l)}) \odot \phi_c(\mathbf{h}_s) \right) + b_{\text{bio}} \right),
\end{equation}
where $\phi_t$ and $\phi_c$ project the token state and the global context into a shared space, $\text{Score}(\cdot)$ is an MLP, and $b_{\text{bio}}$ is a bias derived from the node degree of each gene in $\mathbf{R}$, encoding how strongly a gene is coupled to its regulatory neighbors. The gated residual update then modulates the per-gene contribution of each block:
\begin{equation}
\mathbf{X}^{(l+1)} = \mathbf{X}^{(l)} + g_l \odot \mathcal{F}_l(\mathbf{X}^{(l)}),
\end{equation}
so that informative, tightly-coupled genes receive deeper refinement while stable genes pass through with near-identity updates, which stabilizes optimization across the heterogeneous output space.

\paragraph{Expression Readout.}
After $L$ co-expression blocks, a shared head maps each refined gene token to a
scalar, and a per-gene affine $(\alpha_i,\beta_i)$ rescales it to the target range:
\begin{equation}
\hat{x}_i=\alpha_i\,\mathrm{Head}\!\big(X^{(L)}_i\big)+\beta_i,\qquad
\hat{\mathbf{x}}=[\hat{x}_1,\dots,\hat{x}_G]\in\mathbb{R}^{G},
\end{equation}
in \texttt{log1p} space. This $\hat{\mathbf{x}}=f_\theta(I_s)$ is both the sample
driven by the cellular drifting field and the final inference output, closing the
loop between the generator and the distributional drift objective. For the auxiliary
ZINB term, the same head additionally produces a per-gene
log-dispersion (yielding $\theta_i=\exp(\cdot)>0$) and a per-(spot,\,gene)
zero-inflation logit $\pi_i$, while the ZINB mean is obtained from the point
prediction via $\mu_i=\mathrm{expm1}(\mathrm{ReLU}(\hat{x}_i))$; both are dropped at
inference.
\section{Experiments}
\label{main:experiment}

\subsection{Experimental Settings}
\textbf{Datasets.}
We evaluate on four HEST-1k~\citep{jaume2024hest} datasets: cell-level Xenium Breast Cancer~\cite{janesick2023high} and COAD~\cite{10xgenomics2023xeniumcoad}, and spot-level HER2ST~\cite{andersson2021spatial} and Kidney Visium~\cite{lake2023atlas} (details in Appendix~\ref{app:datasets}).

\textbf{Preprocessing.}
H\&E patches centered on cells or spots are resized to 256$\times$256 for CONCH and 224$\times$224 for UNI2. Gene counts are log1p-normalized to form the drift target, retaining 200–300 HMHVGs per dataset; the corresponding raw integer counts are also kept to evaluate the ZINB likelihood.

\textbf{Evaluation.}
We use spatial 5-fold cross-validation for Xenium datasets and a leave-one-slide-out protocol for spot-level datasets. Metrics include average per-gene Pearson correlation (PCC-10/50/200), MSE, and MAE in $\log_2$ space.

\subsection{Results on Cell-based Datasets}

\begin{table}[H]
\centering
\label{tab:cell_results}
\resizebox{\textwidth}{!}{
\begin{tabular}{l|ccccc|ccccc}
\toprule
& \multicolumn{5}{c|}{Xenium Breast Cancer} & \multicolumn{5}{c}{Xenium COAD} \\
\cmidrule(lr){2-6} \cmidrule(lr){7-11}
Method & PCC-10$\uparrow$ & PCC-50$\uparrow$ & PCC-200$\uparrow$ & MSE$\downarrow$ & MAE$\downarrow$
       & PCC-10$\uparrow$ & PCC-50$\uparrow$ & PCC-200$\uparrow$ & MSE$\downarrow$ & MAE$\downarrow$ \\
\midrule
BLEEP~\cite{xie2023spatially}    & 0.415 & 0.229 & 0.200 & 0.181 & 0.229          & 0.645 & 0.458 & 0.217 & 0.114 & 0.110 \\
M2ORT~\cite{wang2024m2ort}       & 0.515 & 0.409 & 0.192 & \textbf{0.163} & 0.213          & 0.677 & 0.488 & 0.239 & 0.107 & 0.112 \\
TRIPLEX~\cite{chung2024accurate} & 0.539 & 0.427 & 0.218 & 0.166 & 0.222          & 0.681 & 0.489 & 0.212 & \textbf{0.103} & 0.131 \\
STEM~\cite{zhu2025diffusion}     & 0.375 & 0.239 & 0.088 & 0.533 & 0.305          & 0.513 & 0.299 & 0.114 & 0.140 & 0.109 \\
GenAR~\cite{ouyang2025genar}     & 0.361 & 0.201 & 0.070 & 0.531 & 0.332          & 0.541 & 0.309 & 0.110 & 0.148 & 0.142 \\
GHIST~\cite{fu2025ghist}               & 0.458 & 0.364 & 0.172 & 0.194 & 0.226          & 0.592 & 0.427 & 0.134 & 0.135 & \textbf{0.108} \\
sCellST~\cite{chadoutaud2026scellst}       & 0.372 & 0.290 & 0.149 & 0.217 & 0.232          & 0.426 & 0.306 & 0.154 & 0.130 & 0.110 \\
\midrule
\textbf{DriftST (Ours)}          & \textbf{0.587} & \textbf{0.447} & \textbf{0.227} & 0.184 & \textbf{0.212}
                                 & \textbf{0.699} & \textbf{0.518} & \textbf{0.268} & \textbf{0.103} & 0.109 \\
\bottomrule
\end{tabular}
}
\caption{Experimental results on the cell-based datasets (Xenium Breast Cancer and Xenium COAD).
The best results are highlighted in \textbf{bold}.
$\uparrow$ indicates higher is better.}
\end{table}

\textbf{Xenium Breast Cancer.}
DriftST leads on four of five metrics, improving PCC-10/50/200 over the strongest baseline TRIPLEX by 0.048, 0.020, and 0.009 (8.9\% relative on PCC-10) with the lowest MAE (0.212); M2ORT attains the best MSE (0.163) but trails on PCC. Generative methods strong at spot level degrade sharply at cell resolution (e.g., GenAR PCC-10: 0.361), whereas DriftST stays consistent across granularities.

\textbf{Xenium COAD.}
DriftST leads on all three PCC metrics and matches the best MSE (0.103, tied with TRIPLEX), improving PCC-10/50/200 over TRIPLEX by 0.018, 0.029, and 0.056, and still gaining 0.029 on PCC-200 over the strongest long-tail baseline M2ORT (0.239); MAE (0.109) is on par with the best (GHIST, 0.108). The spot-to-cell transfer gap persists, confirming that spot-level models are insufficient for cell-level prediction.

\subsection{Results on Spot-based Datasets}
\begin{table}[H]
\centering
\label{tab:spot_results}
\resizebox{\textwidth}{!}{
\begin{tabular}{l|ccccc|ccccc}
\toprule
& \multicolumn{5}{c|}{HER2ST} & \multicolumn{5}{c}{Kidney Visium} \\
\cmidrule(lr){2-6} \cmidrule(lr){7-11}
Method & PCC-10$\uparrow$ & PCC-50$\uparrow$ & PCC-200$\uparrow$ & MSE$\downarrow$ & MAE$\downarrow$
       & PCC-10$\uparrow$ & PCC-50$\uparrow$ & PCC-200$\uparrow$ & MSE$\downarrow$ & MAE$\downarrow$ \\
\midrule
BLEEP~\cite{xie2023spatially}    & 0.773 & 0.714 & 0.565 & 1.243 & 0.833          & 0.500 & 0.422 & 0.314 & 1.926 & 0.945 \\
M2ORT~\cite{wang2024m2ort}       & 0.810 & 0.759 & 0.660 & 1.151 & 0.820          & 0.494 & 0.447 & 0.318 & 1.785 & 0.925 \\
TRIPLEX~\cite{chung2024accurate} & 0.783 & 0.714 & 0.586 & 1.212 & 0.857          & 0.542 & 0.469 & 0.336 & 1.732 & 0.887 \\
STEM~\cite{zhu2025diffusion}     & 0.831 & 0.770 & 0.625 & 1.199 & 0.787          & 0.567 & 0.483 & 0.322 & 1.832 & 0.997 \\
GenAR~\cite{ouyang2025genar}     & 0.842 & 0.784 & 0.663 & 1.082 & \textbf{0.745} & 0.589 & 0.514 & 0.354 & 1.636 & \textbf{0.871} \\
\midrule
\textbf{DriftST (Ours)}          & \textbf{0.862} & \textbf{0.817} & \textbf{0.757} & \textbf{1.019} & 0.807
                                 & \textbf{0.602} & \textbf{0.532} & \textbf{0.387} & \textbf{1.553} & 0.957 \\
\bottomrule
\end{tabular}
}
\caption{Experimental results on the spot-based datasets (HER2ST and Kidney Visium).
The best results are highlighted in \textbf{bold}.
$\uparrow$ indicates higher is better, $\downarrow$ indicates lower is better.}
\end{table}

\textbf{HER2ST.}
DriftST leads on four of five metrics, improving PCC-10/50/200 over GenAR by 0.020, 0.033, and 0.094. The largest gain is on PCC-200 (14.2\% relative), indicating particular strength on harder-to-model, low-variability genes; MSE also drops from 1.082 to 1.019. GenAR retains an edge on MAE (0.745 vs.\ 0.807), which we attribute to the drift loss optimizing distributional alignment rather than absolute error.

\textbf{Kidney Visium.}
DriftST leads on four metrics, improving PCC-10/50/200 over GenAR by 0.013, 0.018, and 0.033 and reducing MSE from 1.636 to 1.553 (5.1\%). Overall metrics are lower due to severe batch effects and cross-patient heterogeneity under distinct clinical conditions, making generalization an open challenge. GenAR again leads on MAE (0.871 vs.\ 0.957), consistent with HER2ST.

\subsection{Ablation Study}

\begin{table}[H]
\centering
\resizebox{\textwidth}{!}{
\begin{tabular}{l|ccc|ccc|ccc}
\toprule
Model
  & PCC-10$\uparrow$ & PCC-50$\uparrow$ & PCC-200$\uparrow$
  & JSD-20$\downarrow$ & JSD-50$\downarrow$ & JSD-100$\downarrow$
  & SSIM-20$\uparrow$ & SSIM-50$\uparrow$ & SSIM-100$\uparrow$ \\
\midrule
w/o Drift      & 0.691 & 0.509 & 0.264 & 0.504 & 0.519 & 0.509 & 0.535 & 0.468 & 0.411 \\
w/o Co-expr    & 0.683 & 0.498 & 0.240 & 0.418 & 0.479 & 0.467 & 0.599 & 0.532 & 0.498 \\
w/o Gate       & 0.696 & 0.510 & 0.261 & 0.405 & 0.392 & 0.332 & 0.606 & 0.542 & 0.512 \\
\midrule
DriftST (Full) & \textbf{0.699} & \textbf{0.518} & \textbf{0.268} & \textbf{0.401} & \textbf{0.370} & \textbf{0.310} & \textbf{0.644} & \textbf{0.581} & \textbf{0.539} \\
\bottomrule
\end{tabular}
}
\caption{Effect of each component on Xenium COAD. All three variants are ablated from the full DriftST model. $\uparrow$ indicates higher is better, $\downarrow$ indicates lower is better. Best results in \textbf{bold}.}
\label{tab:ablation}
\end{table}

\textbf{Component Ablation.} 
We evaluate three variants on Xenium COAD (Table~\ref{tab:ablation}): w/o Drift (ZINB
loss only), w/o Co-expr (standard self-attention), and w/o Gate. Removing the drift
loss severely degrades distributional alignment (JSD-20: 0.401$\to$0.504, SSIM-20:
0.644$\to$0.535). Dropping the co-expression prior hurts the long tail (PCC-200:
0.268$\to$0.240, JSD-100: 0.310$\to$0.467), proving that biological priors add value
beyond drift loss. Omitting the Gate drops SSIM-20 (0.644$\to$0.606), validating its
role in per-gene spatial fidelity. The full model combines these into synergistic
gains.

\begin{table}[H]
\centering
\resizebox{\textwidth}{!}{
\begin{tabular}{l|ccc|ccc|ccc}
\toprule
Encoder
  & PCC-10$\uparrow$ & PCC-50$\uparrow$ & PCC-200$\uparrow$
  & JSD-20$\downarrow$ & JSD-50$\downarrow$ & JSD-100$\downarrow$
  & SSIM-20$\uparrow$ & SSIM-50$\uparrow$ & SSIM-100$\uparrow$ \\
\midrule
ResNet-18~\citep{he2016deep} & 0.650 & 0.470 & 0.239 & 0.471 & 0.463 & 0.409 & 0.575 & 0.506 & 0.486 \\
CONCH~\citep{lu2024conch}     & 0.697 & 0.511 & 0.263 & 0.405 & 0.381 & 0.320 & 0.641 & 0.577 & 0.539 \\
UNI2~\citep{chen2025uni2}     & 0.691 & 0.507 & 0.261 & 0.412 & 0.393 & 0.335 & 0.637 & 0.572 & 0.533 \\
\midrule
DriftST (Full) & \textbf{0.699} & \textbf{0.518} & \textbf{0.268} & \textbf{0.401} & \textbf{0.370} & \textbf{0.310} & \textbf{0.644} & \textbf{0.581} & \textbf{0.539} \\
\bottomrule
\end{tabular}
}
\caption{Effect of pathology encoder choice on Xenium COAD. $\uparrow$ indicates higher is better, $\downarrow$ indicates lower is better. Best results in \textbf{bold}.}
\label{tab:ablation_encoder}
\end{table}

\textbf{Pathology Encoder Ablation.} Replacing domain-specific foundation models with a general-purpose ResNet-18 drops all metrics substantially (PCC-10: 0.699$\to$0.650), confirming the necessity of specialized pretraining. CONCH slightly outperforms UNI2 due to its vision-language semantic capabilities, while concatenating both yields consistent gains, indicating that morphological and semantic features are complementary. Additional ablation of parameters is provided in Appendix~\ref{app:ablation}

\subsection{Downstream Analysis}

\begin{figure}[h]
\centering
\includegraphics[width=1\linewidth, keepaspectratio]{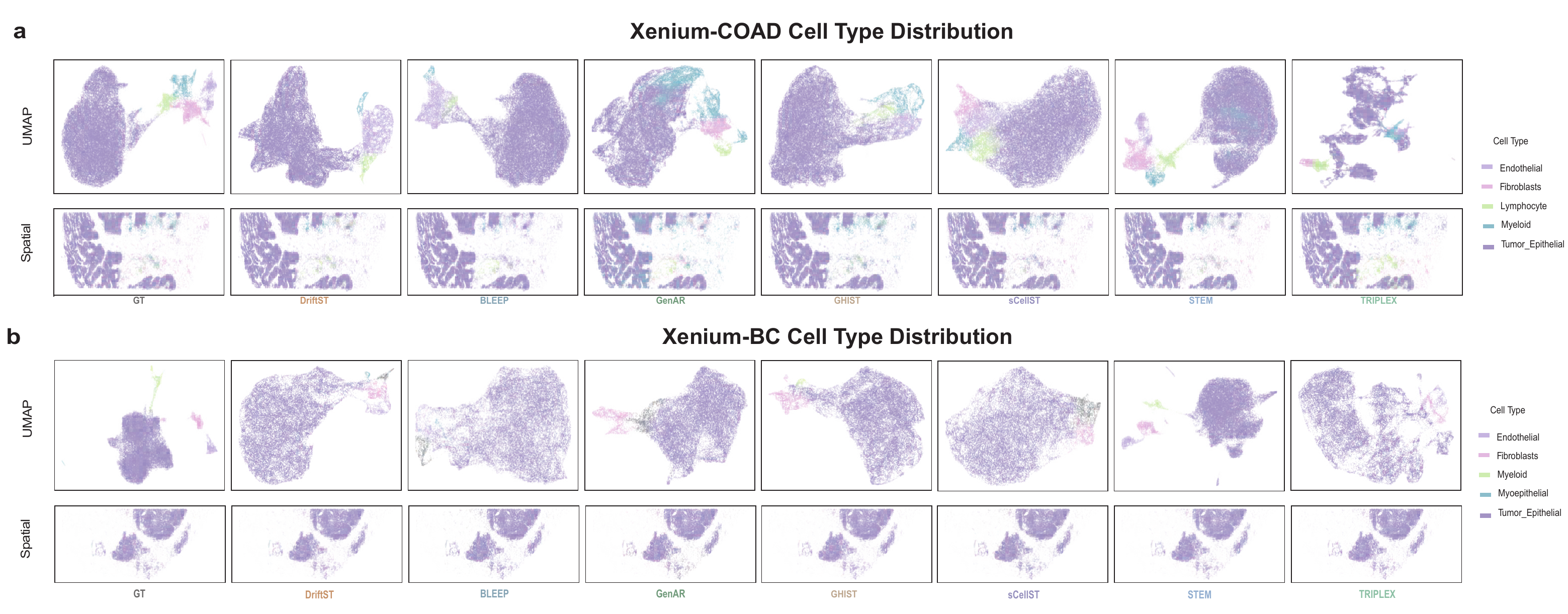}
\caption{\textbf{Cell-type distribution recovered from predicted expression.} Cells are annotated from each method's predicted expression and shown in UMAP (top) and spatial (bottom) coordinates. \textbf{a,} Xenium COAD. \textbf{b,} Xenium BC. Columns: GT and seven methods; colors denote cell types as in the legend.}
\label{fig:xenium}
\end{figure}

\textbf{Cell Level.} We annotate cell types from predicted expressions and compare them in UMAP and spatial spaces (Figure~\ref{fig:xenium}). DriftST faithfully reproduces the main manifold (e.g., Tumor\_Epithelial and Endothelial cells) and recovers minority clusters. Conversely, regression baselines over-smooth and collapse rare cell types into the bulk, while generative baselines distort cluster boundaries, proving DriftST effectively conserves the biological heterogeneity required for downstream cell-type analysis.

\begin{figure}[h]
\centering
\includegraphics[width=1\linewidth, keepaspectratio]{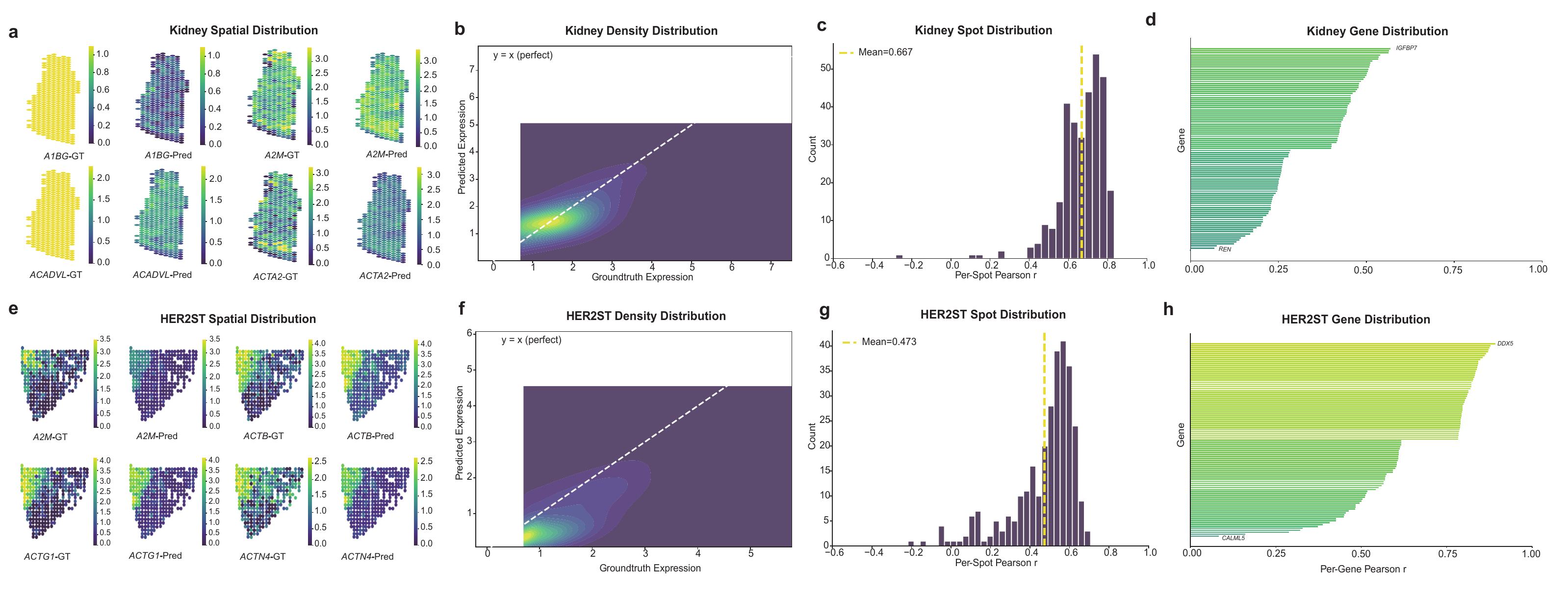}
\caption{\textbf{Prediction quality on spot-level datasets.} \textbf{a--d,} Kidney Visium; \textbf{e--h,} HER2ST. \textbf{a,e,} Spatial maps of GT and predicted expression for representative genes. \textbf{b,f,} Density of predicted vs.\ ground-truth expression, with the $y=x$ line marking perfect prediction. \textbf{c,g,} Histogram of per-spot Pearson $r$ (dashed line: mean). \textbf{d,h,} Per-gene Pearson $r$ sorted from highest to lowest.}
\label{fig:visium}
\end{figure}

\textbf{Spot Level.} DriftST predictions track ground-truth distributions across spatial, density, spot, and gene views (Figure~\ref{fig:visium}). The joint density concentrates along the $y=x$ line, demonstrating well-calibrated magnitudes rather than mean-collapsed outputs. High per-spot correlations (mean $r=0.667$ on Kidney, $0.473$ on HER2ST) and strong recovery of top genes (e.g., IGFBP7, DDX5) confirm that DriftST yields accurate and spatially faithful predictions at spot resolution. Further downstream analysis is provided in Appendix~\ref{app:analysis}
\section{Conclusion}
\label{main:conclusion}

We presented DriftST, a unified framework for inferring spatially resolved gene expression from H\&E histology at both spot- and cell-level resolution. It reformulates the task as Cellular Drifting, a generative model that predicts in a single forward pass via an STransformer with co-expression attention and a gene residual gate, achieving SOTA results across four datasets while preserving the heterogeneity that regression baselines over-smooth. This faithful recovery of spatial variance ensures the reliability of downstream biological discoveries.

DriftST also has limitations that point to future work. Its predictions are restricted to the genes in the training panel rather than the full transcriptome, and cross-patient generalization under heterogeneous tissue conditions and strong batch effects, as observed on Kidney Visium, remains challenging. Extending the framework to transcriptome-wide prediction, jointly modeling multiple slides and patients to learn domain-invariant features, and integrating complementary modalities such as spatial multi-omics are promising directions for broadening its applicability, moving a step closer to practical computational pathology.

\section*{Ethics Statement}
This work uses only publicly available, de-identified spatial transcriptomics
datasets and H\&E images; no new human or animal data were collected. We
complied with dataset licenses and standard citation practices. The models
are released for research use only and are not intended for clinical
decision-making.

\section*{Reproducibility Statement}
We took several steps to support reproducibility. The paper details the
method in Section~\ref{main:method} (architecture, training objective, multi-scale drifting
design), the datasets, gene selection, preprocessing pipeline, and metrics
in Section~\ref{main:experiment}. Appendix~\ref{app:ablation} provides comprehensive ablation studies covering
all key hyperparameters, and Appendix~\ref{app:details} lists complete training and
implementation details (optimizers, batch sizes, scales, hardware).
Source code is available at \url{https://github.com/yyh030806/DriftST}.

\bibliography{iclr2026_conference}
\bibliographystyle{iclr2026_conference}

\newpage
\appendix
\section{Datasets and Evaluation}
\label{app:datasets}

We evaluate DriftST on four spatial transcriptomics datasets spanning two measurement resolutions
(spot-level and cell-level) and three profiling platforms. Table~\ref{tab:dataset_summary}
summarizes the key statistics.

\begin{table}[h]
\centering
\resizebox{\textwidth}{!}{
\begin{tabular}{lccccc}
\toprule
Dataset & Platform & Resolution & Spots/Cells & Genes & Evaluation \\
\midrule
HER2ST & SpatialTranscriptomics & Spot & 13,594 & 300 & Leave-one-slide-out \\
Kidney Visium & 10x Visium & Spot & 25,944 & 200 & Leave-one-slide-out \\
Xenium BC & 10x Xenium In Situ & Cell & 95,273 & 280 & 5-fold spatial CV \\
Xenium COAD & 10x Xenium In Situ & Cell & 161,984 & 280 & 5-fold spatial CV \\
\bottomrule
\end{tabular}
}
\caption{Summary of the four evaluation datasets.}
\label{tab:dataset_summary}
\end{table}

\textbf{HER2ST}~\cite{andersson2021spatial} is a breast cancer dataset sequenced on the
SpatialTranscriptomics platform, containing 36 tissue slides from 8 patients (patients A--D
with six sections each; patients E--H with three consecutive sections each) totalling 13,594
spots at 100\,$\mu$m diameter. The samples include both normal breast tissue and HER2-positive
cancerous regions. We select 300 highly-mean and highly-variable genes (HMHVGs) by ranking
genes according to the product of their mean expression and variance across all spots. We adopt
leave-one-slide-out cross-validation and report results on the held-out test slide SPA138.

\textbf{Kidney Visium}~\cite{lake2023atlas} consists of 23 kidney tissue slides from the Kidney
Precision Medicine Project (KPMP) atlas, profiled on the 10x Visium platform with 55\,$\mu$m
spatial spots. The number of spots per slide ranges from 315 to 4,159, covering three pathological
states (healthy controls, chronic kidney disease, and acute kidney injury) across both cortical
and medullary anatomical regions. Nearly every slide originates from a different patient, making
cross-slide generalization under heterogeneous tissue conditions particularly challenging.
We select 200 HMHVGs using the same ranking criterion as HER2ST.
We adopt the same leave-one-slide-out protocol and report results on the held-out test slide
NCBI697.

\textbf{Xenium Breast Cancer}~\cite{janesick2023high} is a human breast cancer FFPE section
profiled using the 10x Xenium In Situ platform with a 280-gene panel.
After quality control filtering, 95,273 cells are retained. We predict all 280 panel genes.
We partition the single slide into 5 horizontal bands by spatial coordinate and perform 5-fold
cross-validation, reporting average performance across all folds.

\textbf{Xenium COAD}~\cite{10xgenomics2023xeniumcoad} is a human colon
adenocarcinoma FFPE section profiled using the 10x Xenium In Situ platform
with a pre-designed colon gene expression panel supplemented by an add-on panel.
After the same quality control filtering as Xenium Breast Cancer, 161,984 cells
are retained. We select 280 HMHVGs to align with the Xenium Breast Cancer setting.
The same 5-fold spatial cross-validation protocol is applied.
\section{Additional Ablation Study}
\label{app:ablation}

\subsection{Drifting Hyperparameters}
We analyze the sensitivity of the cellular drifting mechanism to its three key
hyperparameters on Xenium COAD: the bandwidth ratio set $\mathcal{R}$, the prediction
bank capacity, and the number of negative samples drawn per update. In addition to the
HMHVG PCC, we report SVG PCC-$k$, the per-gene PCC on the top-$k$ spatially variable
genes (SVG) ranked by Moran's $I$ spatial autocorrelation.

\textbf{Bandwidth Ratio Set $\mathcal{R}$.}
Table~\ref{tab:ablation_rlist} compares single-scale drifting fields against the multi-scale formulation.

\begin{table}[H]
\centering
\begin{tabular}{l|ccc|cc}
\toprule
& \multicolumn{3}{c|}{HMHVG} & \multicolumn{2}{c}{SVG} \\
\cmidrule(lr){2-4} \cmidrule(lr){5-6}
$\mathcal{R}$ & PCC-10$\uparrow$ & PCC-50$\uparrow$ & PCC-200$\uparrow$ & PCC-20$\uparrow$ & PCC-50$\uparrow$ \\
\midrule
$\{0.02\}$              & 0.685 & 0.500 & 0.245 & 0.584 & 0.451 \\
$\{0.05\}$              & 0.677 & 0.495 & 0.241 & 0.573 & 0.439 \\
$\{0.20\}$              & 0.624 & 0.441 & 0.212 & 0.518 & 0.382 \\
$\{0.02, 0.05, 0.2\}$    & \textbf{0.699} & \textbf{0.518} & \textbf{0.268} & \textbf{0.601} & \textbf{0.468} \\
\bottomrule
\end{tabular}
\caption{Ablation on the bandwidth ratio set $\mathcal{R}$.}
\label{tab:ablation_rlist}
\end{table}

Among single-scale settings, smaller bandwidth ratios yield consistently better performance, as fine-grained kernels more precisely resolve local structure in the gene expression space. However, the large-scale kernel ($r = 0.2$) alone performs substantially worse, confirming that coarse interactions are insufficient for accurate distributional transport. The multi-scale combination $\{0.02, 0.05, 0.2\}$ outperforms all individual scales, including the best single scale ($r = 0.02$), demonstrating that aggregating velocity fields across multiple resolutions captures both local precision and global coherence in the drifting dynamics.

\textbf{Prediction Bank Capacity.}
Table~\ref{tab:ablation_banksize} examines the effect of the prediction bank size, which determines the pool of historical negative samples available for constructing the drifting field.

\begin{table}[H]
\centering
\begin{tabular}{c|ccc|cc}
\toprule
& \multicolumn{3}{c|}{HMHVG} & \multicolumn{2}{c}{SVG} \\
\cmidrule(lr){2-4} \cmidrule(lr){5-6}
Bank Size & PCC-10$\uparrow$ & PCC-50$\uparrow$ & PCC-200$\uparrow$ & PCC-20$\uparrow$ & PCC-50$\uparrow$ \\
\midrule
256   & 0.675 & 0.483 & 0.233 & 0.572 & 0.430 \\
1024  & 0.667 & 0.477 & 0.229 & 0.558 & 0.420 \\
4096  & \textbf{0.699} & \textbf{0.518} & \textbf{0.268} & \textbf{0.601} & \textbf{0.468} \\
\bottomrule
\end{tabular}
\caption{Ablation on the prediction bank capacity.}
\label{tab:ablation_banksize}
\end{table}

Performance is non-monotonic at small bank sizes but improves sharply once the bank
is large enough: a bank of 256 and 1024 perform comparably (PCC-10 0.675 vs.\ 0.667),
whereas 4096 yields a pronounced jump to 0.699. A larger bank provides a more
representative and diverse pool of negative samples, enabling the drifting field to
more accurately estimate the pushforward distribution $q_\theta$ and thereby produce
more informative repulsion gradients. The comparable scores at 256 and 1024 suggest
that moderately enlarging the bank without reaching sufficient coverage offers little
benefit, while 4096 crosses a critical threshold where the bank adequately represents
the diversity of generated expression profiles.

\textbf{Bank Sample Size.}
Table~\ref{tab:ablation_samplesize} varies the number of negative samples drawn from the prediction bank for each drifting field computation.

\begin{table}[H]
\centering
\begin{tabular}{c|ccc|cc}
\toprule
& \multicolumn{3}{c|}{HMHVG} & \multicolumn{2}{c}{SVG} \\
\cmidrule(lr){2-4} \cmidrule(lr){5-6}
Sample Size & PCC-10$\uparrow$ & PCC-50$\uparrow$ & PCC-200$\uparrow$ & PCC-20$\uparrow$ & PCC-50$\uparrow$ \\
\midrule
64    & 0.692 & 0.509 & 0.253 & 0.596 & 0.457 \\
256   & 0.681 & 0.490 & 0.231 & 0.576 & 0.437 \\
1024  & \textbf{0.699} & \textbf{0.518} & \textbf{0.268} & \textbf{0.601} & \textbf{0.468} \\
\bottomrule
\end{tabular}
\caption{Ablation on the number of negative samples drawn per update.}
\label{tab:ablation_samplesize}
\end{table}

Using the full bank (1024 samples from a bank of 4096) yields the best results. A larger sample size reduces the variance of the estimated repulsion field, providing a more stable and accurate directional signal for each training step. Interestingly, 64 samples outperform 256, which we attribute to stochastic regularization: very small samples introduce noise that can prevent the drifting field from overfitting to particular negative modes, whereas the intermediate size of 256 offers neither the regularization benefit of high stochasticity nor the low-variance benefit of dense sampling.

\subsection{Gene Residual Gate Hyperparameters}
\label{sec:ablation_gate_hyper}

We examine the sensitivity of the Gene Residual Gate to its regularization hyperparameters on Xenium COAD. The gate is regularized via a sparsity loss weighted by $\lambda_{\text{gate}}$, a target sparsity schedule $(s_{\text{high}}, s_{\text{low}})$ defining the desired gate activation range across layers, and an entropy regularization weight $\lambda_{\text{ent}}$.

\textbf{Gate Regularization Weight $\lambda_{\text{gate}}$.}
Table~\ref{tab:ablation_gateweight} varies the weight of the gate sparsity loss.

\begin{table}[H]
\centering
\begin{tabular}{c|ccc|cc}
\toprule
& \multicolumn{3}{c|}{HMHVG} & \multicolumn{2}{c}{SVG} \\
\cmidrule(lr){2-4} \cmidrule(lr){5-6}
$\lambda_{\text{gate}}$ & PCC-10$\uparrow$ & PCC-50$\uparrow$ & PCC-200$\uparrow$ & PCC-20$\uparrow$ & PCC-50$\uparrow$ \\
\midrule
0.1 & 0.699 & 0.518 & 0.268 & 0.601 & 0.468 \\
0.2 & 0.698 & 0.520 & 0.271 & 0.604 & 0.470 \\
0.5 & 0.699 & 0.519 & 0.271 & 0.607 & 0.473 \\
1.0 & \textbf{0.701} & \textbf{0.522} & \textbf{0.272} & \textbf{0.607} & \textbf{0.472} \\
\bottomrule
\end{tabular}
\caption{Ablation on the gate regularization weight $\lambda_{\text{gate}}$.}
\label{tab:ablation_gateweight}
\end{table}

Performance is remarkably stable across a tenfold range of $\lambda_{\text{gate}}$, with all PCC-10 values falling within a 0.003 band. A slight upward trend is observed at higher weights, indicating that stronger sparsity encouragement marginally improves the gate's ability to differentiate gene complexity.

\textbf{Per-Layer Gene Retention Targets $(r_1, r_2)$.}
Table~\ref{tab:ablation_gatetargets} varies the target gene retention ratio at each transformer layer, where $r_1$ and $r_2$ denote the fraction of genes that remain active (i.e., receive non-negligible gate values) at the first and second layers, respectively. A schedule of $(0.85, 0.50)$, for instance, encourages 85\% of genes to remain active through the first layer while only 50\% continue updating through the second, implementing a progressive funneling of computational resources.

\begin{table}[H]
\centering
\begin{tabular}{c|ccc|cc}
\toprule
& \multicolumn{3}{c|}{HMHVG} & \multicolumn{2}{c}{SVG} \\
\cmidrule(lr){2-4} \cmidrule(lr){5-6}
$(r_1, r_2)$ & PCC-10$\uparrow$ & PCC-50$\uparrow$ & PCC-200$\uparrow$ & PCC-20$\uparrow$ & PCC-50$\uparrow$ \\
\midrule
(0.50, 0.20) & 0.700 & 0.520 & 0.270 & 0.607 & 0.470 \\
(0.70, 0.10) & 0.701 & 0.521 & 0.272 & 0.608 & 0.473 \\
(0.70, 0.20) & \textbf{0.702} & \textbf{0.521} & 0.271 & \textbf{0.610} & \textbf{0.473} \\
(0.85, 0.30) & 0.701 & 0.519 & 0.268 & 0.610 & 0.470 \\
(0.85, 0.40) & 0.700 & 0.520 & 0.270 & 0.606 & 0.471 \\
(0.85, 0.50) & 0.701 & \textbf{0.522} & \textbf{0.272} & 0.607 & 0.472 \\
\bottomrule
\end{tabular}
\caption{Ablation on the per-layer gene retention targets.}
\label{tab:ablation_gatetargets}
\end{table}

All six schedules produce virtually identical performance (PCC-10 range: 0.700--0.702), demonstrating strong robustness to the specific retention targets. Whether the model aggressively prunes early (0.50, 0.20) or retains most genes through the first layer before funneling (0.85, 0.50), the final prediction quality is nearly unchanged. This suggests that the gate learns a biologically grounded allocation of computational depth driven primarily by the intrinsic complexity hierarchy among genes, rather than by the imposed retention schedule.

\textbf{Entropy Regularization Weight $\lambda_{\text{ent}}$.}
Table~\ref{tab:ablation_gateentropy} varies the entropy regularization weight, which encourages gate values to be decisive (near 0 or 1) rather than uniformly intermediate.

\begin{table}[H]
\centering
\begin{tabular}{c|ccc|cc}
\toprule
& \multicolumn{3}{c|}{HMHVG} & \multicolumn{2}{c}{SVG} \\
\cmidrule(lr){2-4} \cmidrule(lr){5-6}
$\lambda_{\text{ent}}$ & PCC-10$\uparrow$ & PCC-50$\uparrow$ & PCC-200$\uparrow$ & PCC-20$\uparrow$ & PCC-50$\uparrow$ \\
\midrule
0    & 0.701 & \textbf{0.523} & \textbf{0.272} & 0.606 & 0.472 \\
0.02 & \textbf{0.702} & 0.521 & 0.271 & \textbf{0.610} & \textbf{0.473} \\
0.05 & 0.702 & 0.519 & 0.269 & 0.609 & 0.472 \\
0.10 & 0.700 & 0.520 & 0.268 & 0.605 & 0.470 \\
\bottomrule
\end{tabular}
\caption{Ablation on the gate entropy regularization weight $\lambda_{\text{ent}}$.}
\label{tab:ablation_gateentropy}
\end{table}

Performance remains stable across all entropy weights, with a marginal preference for $\lambda_{\text{ent}} = 0.02$, which achieves the best PCC on SVG genes. Removing entropy regularization entirely ($\lambda_{\text{ent}} = 0$) yields comparable HMHVG performance but slightly lower SVG prediction accuracy, suggesting that a mild entropy penalty helps produce more decisive gating decisions that translate to slightly improved prediction of spatially variable genes.

\textbf{Summary.} The Gene Residual Gate exhibits strong robustness across all three hyperparameters, with performance varying by less than 0.5\% in PCC-10 across all tested configurations. This stability is a desirable property for practical deployment, as it indicates that the gating mechanism does not require careful hyperparameter tuning to achieve its intended effect of adaptive computational allocation across genes.
\section{Implementation Details and Reproducibility}
\label{app:details}

DriftST is implemented in PyTorch. All experiments use the same set of
hyperparameters across all four datasets.

\textbf{Model.} The encoder consists of a 2-layer residual MLP with hidden dimension
128, followed by 2 Bio-guided Attention layers with 4 attention heads and dropout rate
0.3. Progressive Gene Gating is enabled with gate sparsity weight
$\lambda_{\mathrm{gate}}=0.1$ and entropy regularization weight
$\lambda_{\mathrm{ent}}=0.02$. The number of output genes is 300 for HER2ST, 200 for
Kidney Visium, and 280 for both Xenium datasets.

\textbf{Feature extraction.}
We use frozen UNI2 and CONCH encoders to extract 1,536-d and 512-d
patch embeddings, respectively, which are concatenated into a 2,048-d
input vector per spot or cell. Model weights are publicly available
from their original releases.

\textbf{Training.} We use AdamW with learning rate $3\times10^{-4}$, weight decay
$10^{-4}$, and cosine annealing schedule for 250 epochs with batch size 512. The drift
loss weight is $\alpha=0.15$ with $K=8$ dropout copies per sample. The PredictionBank
has capacity 4{,}096 and 1{,}024 negatives are sampled per batch. The multi-scale
bandwidth ratio set is $\mathcal{R}=\{0.02,0.05,0.2\}$ with step size $\eta=1.0$.
Gradients are clipped to a maximum norm of 1.0.

\textbf{Hardware and runtime.}
All experiments are conducted on a single NVIDIA RTX 4090 GPU.
On the largest dataset (Xenium COAD, 161,984 cells), one fold takes
approximately 7 hours (250 epochs, $\sim$1.7 min/epoch), with the best
checkpoint typically reached around epoch 100. For cell-level datasets
with 5-fold cross-validation, total training time is approximately
35 hours per dataset.

\textbf{Reproducibility.}
Source code and preprocessed data are publicly available at \url{https://github.com/yyh030806/DriftST}.
\section{Training and Inference Algorithm}
\label{app:algorithm}

Algorithm summarizes the DriftST training procedure.

\textbf{Training.}
Before training begins, the PredictionBank is populated by running
$K$ stochastic forward passes over the training set to ensure a diverse
negative pool is available from the first iteration.
At each training step, each input is forwarded $K$ times through the model
with dropout active, producing $K$ stochastic predictions per spot.
Only the first forward pass retains gradients; the remaining $K{-}1$
passes are detached to reduce memory cost.
After each batch, all $B \times K$ predictions are enqueued into the
PredictionBank as future negative samples.

\textbf{Inference.}
At test time, DriftST performs a single deterministic forward pass
(with dropout disabled), directly mapping the concatenated image features
to predicted gene expression.
Neither the PredictionBank nor iterative decoding is required,
making inference substantially simpler than existing generative
approaches that rely on multi-step iterative decoding or denoising.

\begin{algorithm}
\caption{DriftST Training}
\begin{algorithmic}[1]
\Require Dataset $\mathcal{D}=\{(\mathbf{I}_i,\mathbf{x}_i,\mathbf{x}_i^{\mathrm{cnt}})\}$,
         bank capacity $M$, dropout copies $K$, negative size $N$,
         bandwidth ratio set $\mathcal{R}$
\State Initialize model $f_\theta$, PredictionBank $\mathcal{B}\leftarrow\emptyset$
\State $\textsc{WarmUpBank}(f_\theta,\mathcal{D},\mathcal{B},K)$
       \Comment{pre-fill bank with $K$-fold dropout predictions}
\For{epoch $=1,\dots,E$}
  \For{each mini-batch $\{(\mathbf{I}_i,\mathbf{x}_i,\mathbf{x}_i^{\mathrm{cnt}})\}_{i=1}^{B}$}
    \State $\hat{\mathbf{x}}_i^{(1)},\dots,\hat{\mathbf{x}}_i^{(K)}\leftarrow K$ stochastic forward passes of $f_\theta(\mathbf{I}_i)$
           \Comment{only $\hat{\mathbf{x}}^{(1)}$ retains gradient; $f_\theta$ also emits $\theta,\pi$}
    \State $\mu_i\leftarrow\mathrm{expm1}(\mathrm{ReLU}(\hat{\mathbf{x}}_i^{(1)}))$;\quad
           $\mathcal{L}_{\mathrm{ZINB}}\leftarrow\mathrm{ZINB}(\mathbf{x}_i^{\mathrm{cnt}},\mu_i,\theta,\pi)$
    \State $\mathbf{n}_1,\dots,\mathbf{n}_N\leftarrow\textsc{Sample}(\mathcal{B},N)$
           \Comment{historical negatives from the bank}
    \For{each $r\in\mathcal{R}$}
      \State $A^{+}_r,A^{-}_r\leftarrow$ split-form softmax over positive $\mathbf{x}_i$ and negatives $\{\hat{\mathbf{x}}_i^{(k)}\}_{k}\cup\{\mathbf{n}_j\}_{j}$ \Comment{self excluded}
      \State $\mathbf{v}_r\leftarrow A^{+}_r\,\mathbf{x}_i-A^{-}_r\,[\hat{\mathbf{x}}_i^{(k)};\mathbf{n}_j]$
    \EndFor
    \State $g\leftarrow\hat{\mathbf{x}}_i^{(1)}+\dfrac{\eta}{|\mathcal{R}|}\sum_{r}\mathbf{v}_r$
           \Comment{drift goal (detached)}
    \State $\mathcal{L}_{\mathrm{drift}}\leftarrow\|\hat{\mathbf{x}}_i^{(1)}-g\|^2$
    \State $\mathcal{L}\leftarrow\alpha\,\mathcal{L}_{\mathrm{drift}}+(1-\alpha)\,\mathcal{L}_{\mathrm{ZINB}}+\lambda_{\mathrm{gate}}\,\mathcal{L}_{\mathrm{gate}}$
    \State update $\theta$ by $\nabla_\theta\mathcal{L}$
    \State $\mathcal{B}.\textsc{enqueue}(\hat{\mathbf{x}}_i^{(1)},\dots,\hat{\mathbf{x}}_i^{(K)})$
           \Comment{ring buffer update}
  \EndFor
\EndFor
\end{algorithmic}
\end{algorithm}
\section{Further Downstream Analysis}
\label{app:analysis}

\begin{figure}[h]
\centering
\includegraphics[width=1\linewidth, keepaspectratio]{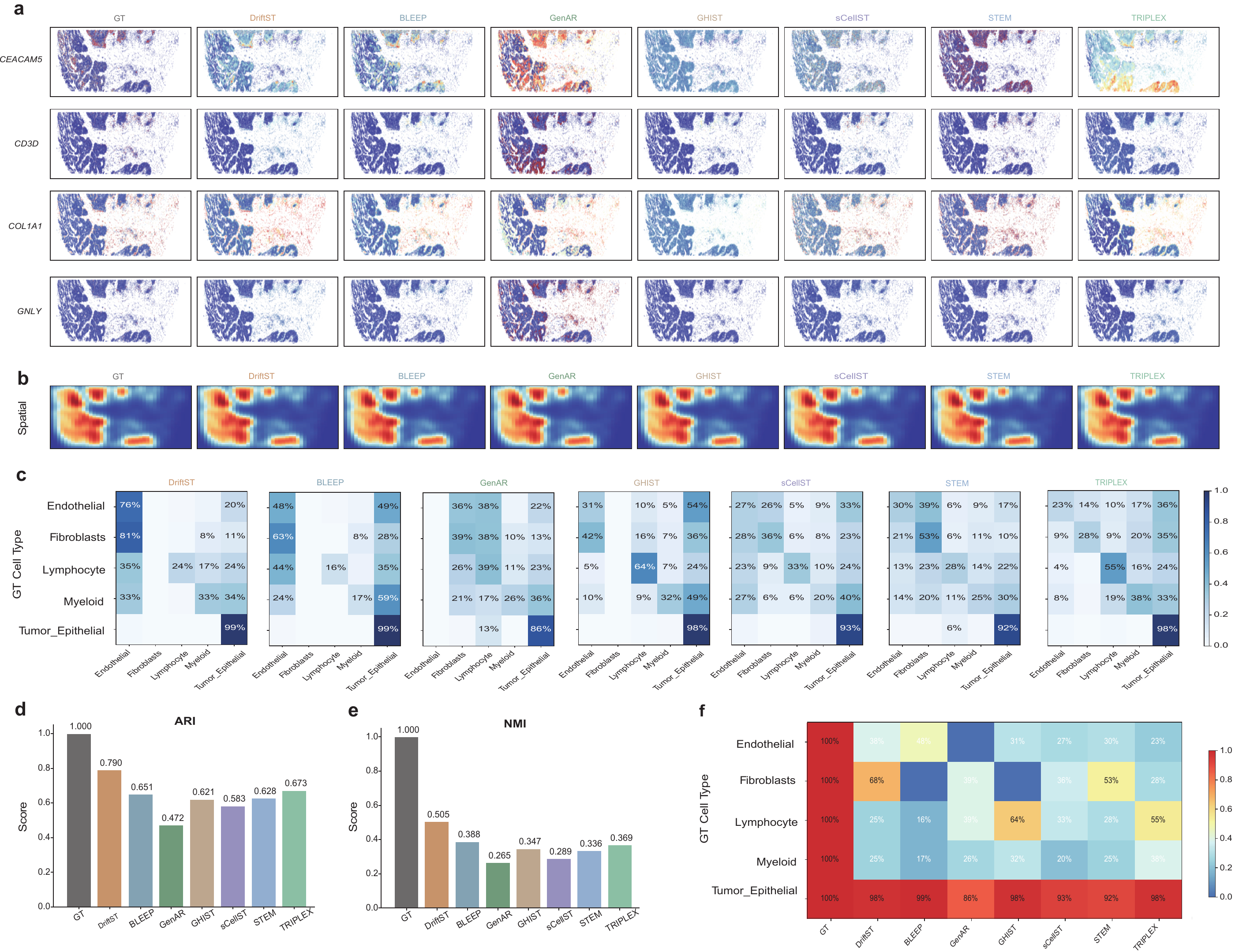}
\caption{\textbf{Downstream analysis on the Xenium colorectal adenocarcinoma (COAD) dataset.} \textbf{a,} Spatial expression of representative lineage markers (CEACAM5, CD3D, COL1A1, GNLY) for the ground truth (GT) and predictions from DriftST and six baselines. \textbf{b,} Tissue-level spatial expression maps, comparing the reconstructed spatial structure against GT. \textbf{c,} Row-normalized confusion matrices of cell-type assignment, obtained by clustering each method's predicted profiles and aligning the clusters to GT annotations (rows: GT cell types). \textbf{d,e,} Adjusted Rand Index (ARI) and Normalized Mutual Information (NMI) between the predicted and GT clusterings. \textbf{f,} Per-cell-type recovery (diagonal recall) summarized across all evaluated methods.}\label{fig:xenium_coad}
\end{figure}

\begin{figure}[h]
\centering
\includegraphics[width=1\linewidth, keepaspectratio]{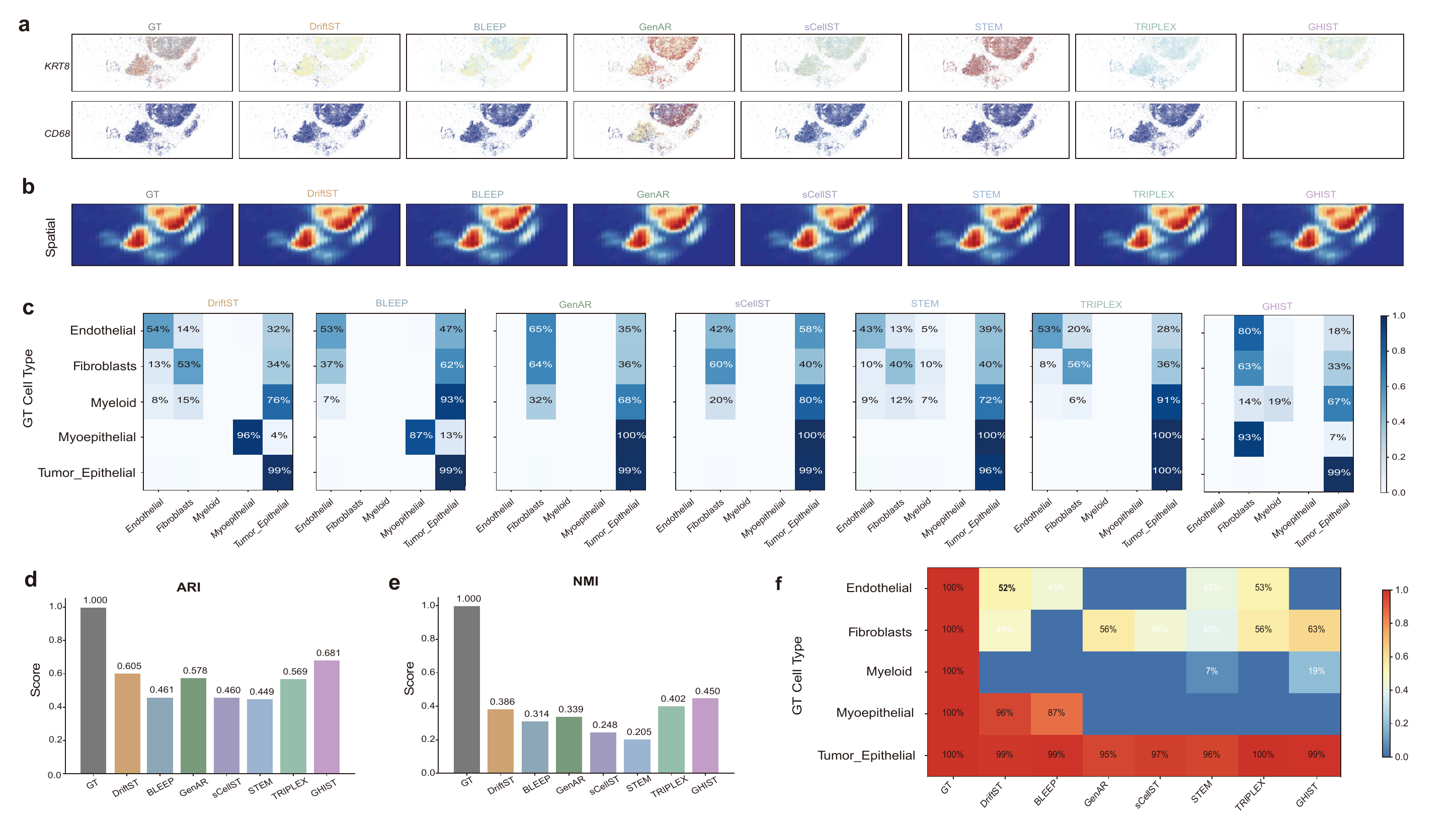}
\caption{\textbf{Downstream analysis on the Xenium breast cancer (BC) dataset.} \textbf{a,} Expression of representative lineage markers (KRT8, CD68) shown on the cell embedding for GT and all methods. \textbf{b,} Tissue-level spatial expression maps. \textbf{c,} Row-normalized cell-type confusion matrices from clustering the predicted profiles (rows: GT cell types, including the breast-specific Myoepithelial compartment). \textbf{d,e,} ARI and NMI between the predicted and GT clusterings. \textbf{f,} Per-cell-type recovery summarized across methods.}\label{fig:xenium_bc}
\end{figure}

To assess whether the profiles generated by DriftST support biologically meaningful downstream analysis rather than merely minimizing per-gene error, we evaluate two single-cell resolution Xenium sections:a colorectal adenocarcinoma sample (COAD, Figure \ref{fig:xenium_coad}) and a breast cancer sample (BC, Figure \ref{fig:xenium_bc}) . For each, we ask whether the predictions preserve marker gene localization, reproduce tissue-level spatial structure, and retain enough cell-state information to recover the underlying cellular composition through unsupervised clustering.

\textbf{Marker gene and spatial fidelity.} Figures \ref{fig:xenium_coad}a and \ref{fig:xenium_bc}a visualize representative lineage markers spanning the major compartments: epithelial (KRT8, CEACAM5), myeloid (CD68), fibroblast (COL1A1), and lymphoid (CD3D, GNLY). DriftST reproduces the GT pattern of each marker, keeping high-expression regions confined to their corresponding compartments and recovering the sharp boundary between the malignant epithelium and the surrounding stroma. Several baselines either blur these boundaries or over-amplify the signal; GenAR in particular produces diffuse, saturated maps in which marker specificity is largely lost. Panel b confirms this at the tissue level: the spatial expression map from DriftST is the closest visual match to GT, preserving both the position and the relative intensity of the high-expression foci, whereas competing methods tend to smooth out or displace these structures.

\textbf{Cell-type composition recovery.} To probe how well each prediction retains discrete cell identities, we cluster the generated profiles and align the resulting clusters to the GT annotations, reporting row-normalized confusion matrices (Figures \ref{fig:xenium_coad}c and \ref{fig:xenium_bc}c). The dominant malignant compartment is recovered almost perfectly by every method: Tumor\_Epithelial recall remains at $99\%$ on BC and ranges from $86\%$ to $99\%$ on COAD, indicating that the transcriptomic phenotype of tumor cells is readily reconstructed from morphology. The discriminating factor lies in the minority compartments. On BC, DriftST is the only method besides BLEEP to recover the Myoepithelial layer ($96\%$ versus $87\%$), whereas GenAR, sCellST, STEM, and TRIPLEX collapse it entirely into Tumor\_Epithelial ($100\%$ misassignment), erasing a clinically important basal population. On COAD, DriftST attains the highest Endothelial recall ($76\%$) together with near-perfect Tumor\_Epithelial recovery ($99\%$), although it tends to absorb part of the Fibroblast population into the Endothelial cluster. Rare immune cells remain the hardest target for all methods: Myeloid (both datasets) and Lymphocyte (COAD) are frequently routed into the tumor or stromal clusters, with only GHIST ($64\%$) and TRIPLEX ($55\%$) recovering an appreciable fraction of lymphocytes. Panel f summarizes this structure across methods, showing a near-uniformly high tumor row alongside progressively harder rare-cell rows.

\textbf{Clustering concordance.} We quantify the global agreement between the predicted and GT clusterings using ARI and NMI (Figures \ref{fig:xenium_coad}d,e and \ref{fig:xenium_bc}d,e). On COAD, DriftST achieves the highest scores among all methods on both metrics (ARI $0.790$, NMI $0.505$), clearly ahead of the next-best TRIPLEX (ARI $0.673$) and BLEEP (NMI $0.388$), reflecting a substantially cleaner separation of cellular states. On BC, the ranking is tighter: DriftST stays competitive (ARI $0.605$, NMI $0.386$), ranking second in ARI behind GHIST ($0.681$) and within the leading group on NMI, where GHIST ($0.450$) and TRIPLEX ($0.402$) are strongest. This contrast is consistent with the confusion structure of the two tissues. BC is dominated by a few well-separated compartments on which several methods perform comparably, whereas COAD presents a more granular immune and stromal landscape, on which the distributional modeling of DriftST yields a clearer advantage.

\textbf{Summary.} Across both tissues, DriftST does not merely fit per-gene values but generates expression profiles that retain cell-state structure usable for clustering, marker-based annotation, and spatial interpretation. It recovers compartments that competing methods erase, such as the breast Myoepithelial layer, and delivers the strongest clustering concordance on the more heterogeneous COAD section. The results also identify rare immune populations as a limitation shared by all current methods, marking a clear direction for future work.
\section{Large Language Model Usage}

Large Language Models were used as general-purpose writing assistance tools to improve the grammar,
clarity, and organization of the manuscript. The core research contributions, methodology, experimental
design, and scientific insights are entirely original work by the authors.
\end{document}